\documentclass[twocolumn]{article}
\usepackage{amsmath}
\usepackage{graphicx}

\usepackage{amsmath}
\usepackage{amsfonts}
\usepackage{amssymb}
\usepackage{url}
\usepackage{bm}
\usepackage{bbm}
\usepackage{float}
\usepackage{algorithm}
\usepackage{algpseudocode}
\usepackage{multirow}
\usepackage{bbm}
\usepackage[table,xcdraw]{xcolor}

\makeatletter
\newcommand{\algmargin}{\the\ALG@thistlm}
\makeatother
\newlength{\whilewidth}
\algdef{SE}[parWHILE]{parWhile}{EndparWhile}[1]
{\parbox[t]{\dimexpr\linewidth-\algmargin}{%
		\hangindent\whilewidth\strut\algorithmicwhile\ #1\ \algorithmicdo\strut}}{\algorithmicend\ \algorithmicwhile}%
\algnewcommand{\parState}[1]{\State%
	\parbox[t]{\dimexpr\linewidth-\algmargin}{\strut #1\strut}}

\def\L{{\cal L}}
\DeclareMathOperator*{\argmin}{argmin}

\title{VQVAE Unsupervised Unit Discovery \\
	and Multi-scale Code2Spec Inverter\\for Zerospeech Challenge 2019}
\author{Andros Tjandra$^{1,2}$, Berrak Sisman$^{3}$, Mingyang Zhang$^{3}$, Sakriani Sakti$^{1,2}$, \\Haizhou Li$^{3}$, Satoshi Nakamura$^{1,2}$ \\
\\ $^1$Nara Institute of Science and Technology, Japan\\
$^2$RIKEN, Center for Advanced Intelligence Project AIP, Japan, \\ 
$^3$Department of Electrical and Computer Engineering, National University of Singapore, Singapore\\ \texttt{\{andros.tjandra.ai6\}@is.naist.jp}
}
\date{}

\begin{document}
	
	\maketitle
	
	\begin{abstract}
		We describe our submitted system for the ZeroSpeech Challenge 2019. The current challenge theme addresses the difficulty of constructing a speech synthesizer without any text or phonetic labels and requires a system that can (1) discover subword units in an unsupervised way, and (2) synthesize the speech with a target speaker's voice. Moreover, the system should also balance the discrimination score ABX, the bit-rate compression rate, and the naturalness and the intelligibility of the constructed voice. To tackle these problems and achieve the best trade-off, we utilize a vector quantized variational autoencoder (VQ-VAE) and a multi-scale codebook-to-spectrogram (Code2Spec) inverter trained by mean square error and adversarial loss. The VQ-VAE extracts the speech to a latent space, forces itself to map it into the nearest codebook and produces compressed representation. Next, the inverter generates a magnitude spectrogram to the target voice, given the codebook vectors from VQ-VAE. In our experiments, we also investigated several other clustering algorithms, including K-Means and GMM, and compared them with the VQ-VAE result on ABX scores and bit rates. Our proposed approach significantly improved the intelligibility (in CER), the MOS, and discrimination ABX scores compared to the official ZeroSpeech 2019 baseline or even the topline.
	\end{abstract}
	
	\section{Introduction}
	 
	Current spoken language technologies only cover about two percent of the world's languages.
	This is because most groundworks require a large amount of paired data resources, including a sizeable 
	collection of spoken audio data and corresponding text transcription. On the other hand, most of the world's languages are severely
	under-resourced, some of which even lack a written form. Zero resource speech research is an extreme case from low-resourced approaches that learn the elements of a language solely from untranscribed raw audio data. This completely unsupervised technique
	attempts to mimic the early language acquisition of humans. The zero resource speech challenge
	(ZeroSpeech) \cite{versteegh2015zero, dunbar2017zero, zs2019} is directly addressing this issue and offers participants the opportunity to advance the state-of-the-art in the core tasks of zero resource speech technology.
	
	In ZeroSpeech 2015 and 2017, the goal was to discover an appropriate speech representation of the
	underlying language of a dataset \cite{versteegh2015zero, dunbar2017zero}. The ZeroSpeech 2019 \cite{zs2019} challenge confronts the problem of constructing a speech synthesizer without any text or phonetic labels: TTS without T.
	The task requires the full system not only to discover subword units in an unsupervised way but also
	to re-synthesize the speech with a same content  to a different target speaker. It includes both ASR and TTS components. In this paper, we describe our submitted system for the ZeroSpeech Challenge 2019 and focus on constructing end-to-end systems.
	
	The top performances in discovering speech representation in ZeroSpeech 2015 and 2017 are dominated by
	a Bayesian non-parametric approach with unsupervised cluster speech features using a Dirichlet process Gaussian mixture model (DPGMM) \cite{chen2015parallel, heck2017feature}. However, the DPGMM model is too sensitive to acoustic variations and often produces too many subword units and a relatively high-dimensional posteriogram, which implies high computational cost for learning and inference as well as more tendencies for overfitting \cite{wuoptimizing}. Therefore it is difficult to synthesize speech waveform from the resulting DPGMM-based acoustic units.
	
\begin{figure}[]
		\centering
		\includegraphics[width=0.6\linewidth]{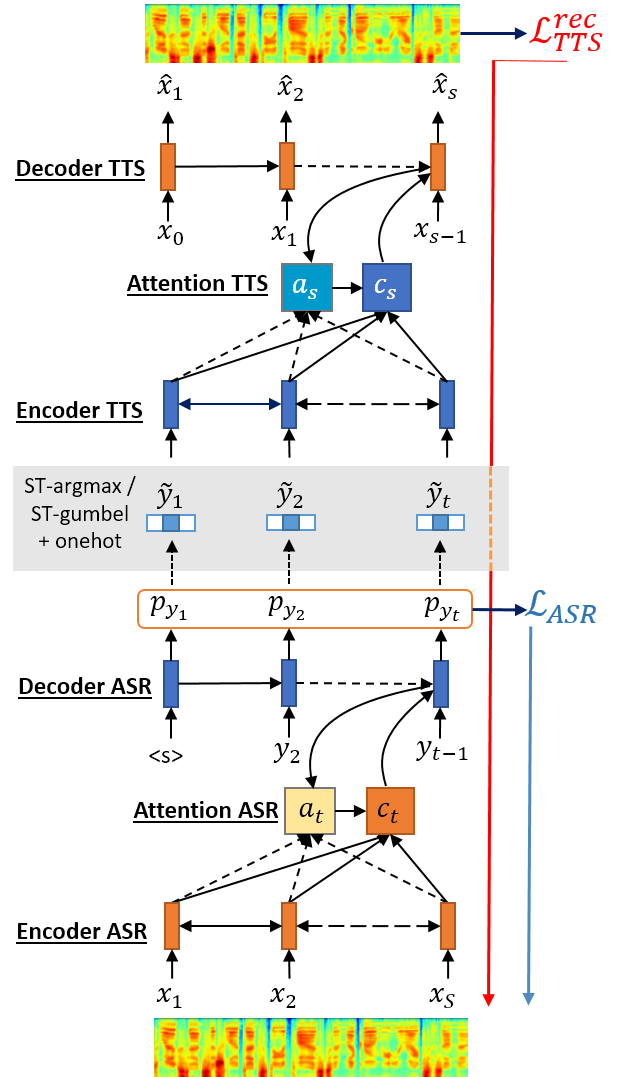}
		
		\caption{Speech chain model consists of an ASR and a TTS module. Given speech features $\mathbf{x}$, the ASR generates a transcription. The TTS reconstructs the speech features based on the transcription and calculates the reconstruction loss $\L^{rec}_{tts}$.}
		\label{fig:speech_chain}
		
	\end{figure}

	To tackle these problems and achieve the best trade-off, an optimization method is required to balance and improve both components. Recently, Tjandra et al. \cite{tjandra2017listening, tjandra2018machine, tjandra2019stestimator} proposed a machine speech chain (see Figure~\ref{fig:speech_chain}) that enables ASR and TTS to assist each other when they receive unpaired data by allowing them to infer the missing pair and optimize both models with reconstruction loss. However, since the architecture is based on an attention-based sequence-to-sequence framework that transforms from a dynamic-length input into a dynamic-length output without decoding at the frame-level (one symbol per frame), it is less suitable for this challenge.
	
	Inspired by a similar idea, we propose to utilize a frame-based vector quantized variational autoencoder (VQ-VAE) \cite{van2017neural} and a multi-scale codebook-to-spectrogram (Code2Spec) inverter trained by mean square error (MSE) and adversarial loss. VQ-VAE extracts the speech to a latent space and forces itself to map onto the nearest codebook, leading to compressed representation. Next, the inverter generates a magnitude spectrogram to the target voice, given the codebook vector  from VQ-VAE. In our experiments, we also investigate other clustering algorithms such as K-Means and GMM and compare them with the VQ-VAE result on ABX scores and bit rate.
	 
	\section{Vector Quantized Variational Autoencoder (VQ-VAE)}
	 
	A vector quantized variational autoencoder (VQ-VAE) \cite{van2017neural} is a variant of variational autoencoder architecture. It has several differences compared to a standard autoencoder or a variational autoencoder \cite{kingma2013auto} (VAE). First, the encoder generates discrete latent variables instead of continuous latent variables to represent the input data. Second, instead of one-to-one mapping between the input data and the latent variables, VQ-VAE forces the latent variables to be represented by the closest codebook vector.
	\begin{figure}[]
		\centering
		\includegraphics[width=0.8\linewidth]{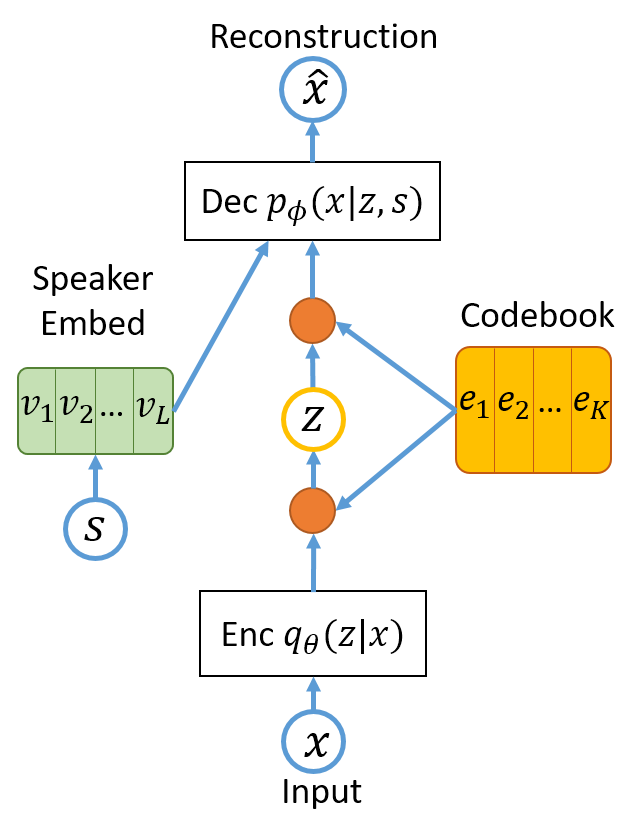}
		 
		\caption{Conditional VQ-VAEs consist of four main modules: encoder $q_\theta(z|x)$, decoder $p_\phi(x|z,s)$, codebooks $E =[e_1,..,e_K]$, and speaker embedding $V = [v_1,..,v_L]$.}
		\label{fig:vqvae}
		 
	\end{figure}

	Figure~\ref{fig:vqvae} illustrates the encoding and decoding processes from the conditional VQ-VAE model. Here $x$ is the input data, $s \in \{1,..,L\}$ is the speaker ID that is related to $x$, $z \in \{1,..,K\}$ is a discrete latent variable, and $\hat{x}$ is the reconstructed input. Encoder $q_\theta(z|x)$ and decoder $p_\phi(x|z,s)$ can be represented by any differentiable transformation (e.g., linear, convolution, recurrent layer) parameterized by $\{ \phi, \theta \}$. Codebook $E = [e_1, e_2,..,e_K] \in \mathbb{R}^{K \times D_e}$ is a collection of $K$ continuous codebook vectors with $D_e$ dimensions. Speaker embedding $V = [v_1, v_2, ..., v_L] \in \mathbb{R}^{L \times D_v}$ is speaker embedding to map speaker ID $s$ into a continuous representation. In the encoding step, encoder $q_\theta(z|x)$ projects input $x$ into continuous representation $\hat{z} \in \mathbb{R}^{D_e}$. Posterior distributions $q_\theta(z|x)$ are generated by a discretization process:
	\begin{align}
	q_{\theta}(z=c|x) &= \begin{cases}
	1 \quad \text{if } \, c=\argmin_{i} \| \hat{z} - e_i \|_2\\
	0 \quad \text{else }
	\end{cases} \\
	e_c &= \sum_{i=1}^{K} q_\theta(z=i | x) \, e_i.
	\end{align}
	In the discretization process, we choose closest codebook vector $e_c$ based on the index of the closest distance (e.g., L2-norm distance) from continuous representation $\hat{z}$. To decode the data, we use codebook $e_c$ and speaker embedding $v_s$ and feed both into decoder $p_\phi(x|z, s) = p_\phi(x|e_c, v_s)$ to reconstruct original data $\hat{x}$.
	
	%In the training process, VAE are optimized by minimize the evidence lower bound (ELBO).
	In VQ-VAE, we formulate the training objective:
	\begin{equation}
	\mathcal{L}_{VQ} = -\log p_\phi(x|z, s) + \| \text{sg}(\hat{z}) - e_c \|_2^2 + \gamma \| \hat{z} - \text{sg}(e_c) \|_2^2,
	\end{equation} where function $\text{sg}(\cdot)$ stops the gradient, defined as:
	\begin{equation}
	x = sg(x) ; \quad
	\frac{\partial \,\text{sg}(x)}{\partial \,x} = 0.
	\end{equation}
	There are three terms in loss $\mathcal{L}_{VQ}$. The first is a negative log-likelihood that resembles  a reconstruction loss and optimizes the encoder and decoder parameters. The second optimizes codebook vectors $E$, named codebook loss. The third forces the encoder to generate a representation near the codebook, called commitment loss. Coefficient $\gamma$ is used to scale the commitment loss.

	\section{Codebook-to-Spectrogram Inverter}
	 
	The codebook-to-spectrogram (Code2Spec) inverter is a module that reconstructs the speech signal representation (e.g., linear magnitude spectrogram) $M = [m[1], m[2], ..., m[T_s]] \in \mathbb{R}^{T_s \times D_m }$, given a sequence of codebook $[e[1],e[2],...,e[T_z]] \in \mathbb{R}^{T_z \times D_e}$.
	\begin{figure}[]
		\centering
		\includegraphics[width=0.9\linewidth]{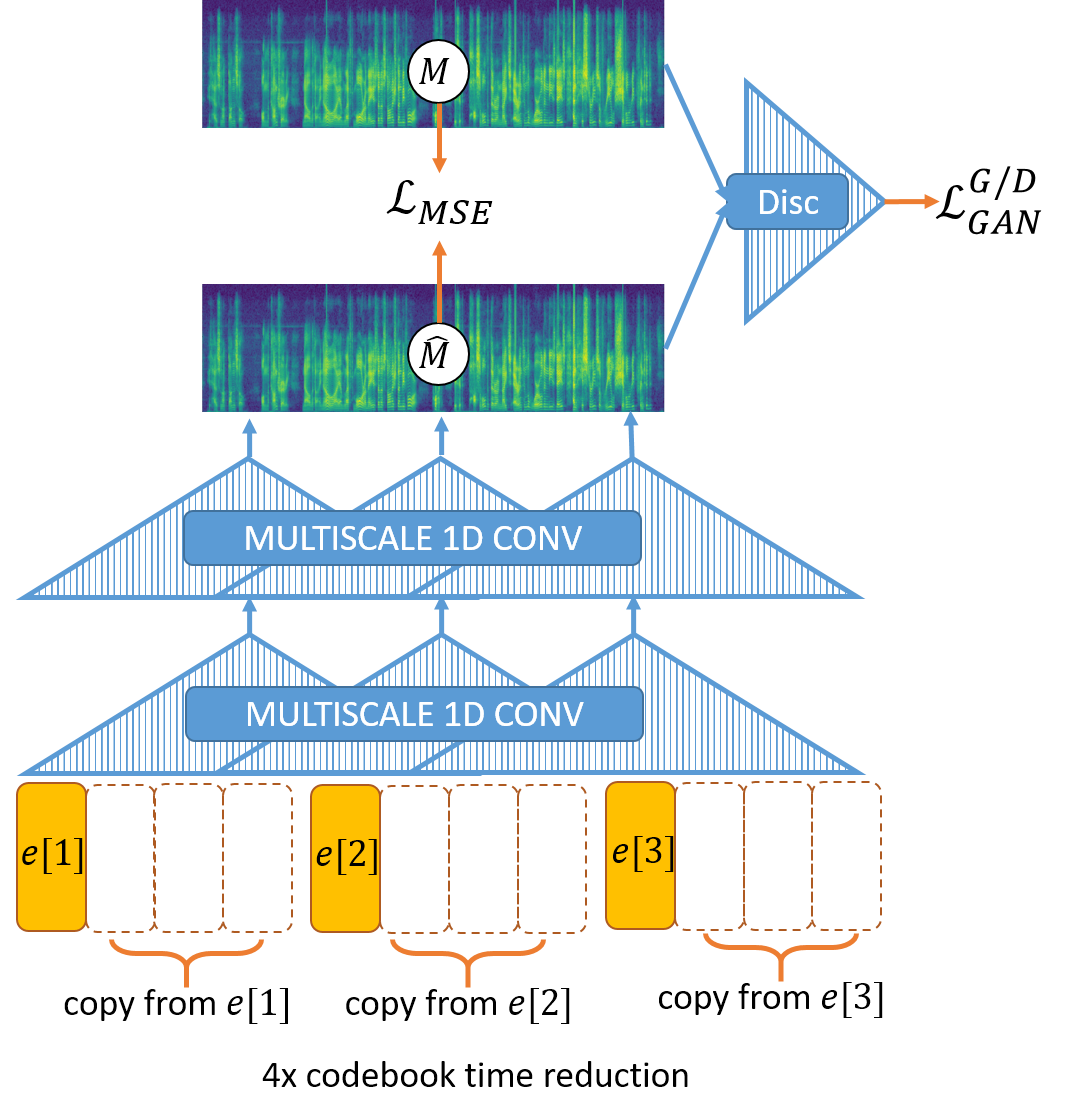}
		  
		\caption{Code-to-speech inverter: given a sequence of codebook $\left[e[1], e[2],.., e[T_z]\right]$, we duplicate each codebook based on compression ratio $r=4$ and apply multiple layers of multi-scale 1D convolution + LeakyReLU activation function to predict the target voice linear spectrogram $\hat{M}$.}
		\label{fig:inverter}
		 
	\end{figure}
	
	In Fig.~\ref{fig:inverter}, we illustrate our code-to-speech inverter model. The length of codebook sequence $T_z$ might be shorter than $T_s$, depending on the VQ-VAE encoder $q_\theta(z|x)$ model. Therefore, for an identical length between the codebook and speech representation sequences, we need to copy $r = T_s / T_z$ times for each codebook $e[t]; \, \forall t \in [1..T_z]$. Later, duplicated codebook sequences $[e[1], e[1], .., e[T_z], e[T_z]] \in \mathbb{R}^{T_s \times D_e}$ are given to the inverter that consists of multiple layers of multi-scale 1D convolution, batch-normalization \cite{ioffe2015batch}, and LeakyReLU \cite{maas2013rectifier} non-linearity. In addition to the inverter, we also have a discriminator module. The discriminator predicts whether the given spectrogram is real data or is generated by the inverter, which generates a realistic spectrogram to deceive the discriminator \cite{goodfellow2014generative, saito2018gan, kaneko2017generative}.
	The Code2Spec inverter has several training objectives:
	\begin{align}
	\hat{M} &= \text{Code2Spec}([e[1], e[1], .., e[T_z], e[T_z]]) \\
	\mathcal{L}_{MSE} &= \|M - \hat{M}\|_2^2 \\
	\mathcal{L}_{GAN}^{G} &= \begin{cases}
	-\text{Disc}(\hat{M}) & \text{WGAN \cite{arjovsky2017wasserstein}} \\
	(\text{Disc}(\hat{M})-1)^2 & \text{LSGAN \cite{mao2017least}}
	\end{cases} \\
	\mathcal{L}_{GAN}^{D} &= \begin{cases}
	\text{Disc}(\hat{M}) - \text{Disc}(M) & \text{WGAN} \\
	\text{Disc}(\hat{M})^2 + (\text{Disc}(M)-1)^2 & \text{LSGAN}
	\end{cases}
	\end{align}
	After we define the multiple objectives for training, we update each module parameter $\theta_{C2S}$ and $\theta_{Disc}$ with the following equation:
	\begin{eqnarray}
	\theta_{C2S} &=& \text{Optim}(\theta_{C2S}, \nabla_{\theta_{C2S}}(\alpha \L_{MSE} + \beta \L_{GAN}^{G})) \\
	\theta_{Disc} &=& \text{Optim}(\theta_{Disc}, \nabla_{\theta_{Dssc}}(\L_{GAN}^D)),
	\end{eqnarray} where Optim($\cdot, \cdot$) is a gradient optimization function (e.g., SGD, Adam \cite{kingma2014adam}), $\alpha$  and $\beta$ is the coefficient to balance the loss between the MSE and the adversarial loss.
	In the inference stage, given the predicted linear magnitude spectrogram $\hat{M}$, we reconstruct the missing phase spectrogram with the Griffin-Lim algorithm \cite{griffin1984signal} and applied the inverse short-term Fourier transform (STFT) to generate the waveform.

	\section{Experiment}
	In this section, we describe the feature extraction, the preliminary models, and our proposed models for this challenge.
	All of the results were evaluated using \texttt{evaluate.sh} from the English test set.
	 
	\subsection{Experimental Set-up}
	There are two datasets for two languages, English data for the development dataset, and a surprise Austronesian language for the test dataset.
	Each language dataset contains subset datasets: (1) a Voice Dataset for speech synthesis, (2) a Unit Discovery Dataset, (3) an Optional Parallel Dataset from the target voice to another speaker voice, and (4) a Test Dataset. The source corpora of the surprise language are describe here \cite{indotts, indoasr}, and further details can be found here \cite{zs2019}. In this work, we only use (1)-(2) for training and (4) for testing.
	
	For the speech input, we experimented with several feature types, such as Mel-spectrogram (80 dimensions, 25-ms window size, 10-ms time-steps) and MFCC (13 dimensions$+\Delta+\Delta^2$ (total=39 dimensions), 25-ms window size, 10-ms time-steps). Both MFCC and Mel-spectrogram are generated by the Librosa package \cite{mcfee2015librosa}.

	\subsection{Official baseline and topline model}
	ZeroSpeech 2019 provides official baselines and toplines. The baseline consists of a pipeline with a simple acoustic unit discovery system based on DPGMM and a speech synthesizer based on Merlin, and the topline uses gold phoneme transcription to train a phoneme-based ASR system with Kaldi and a phoneme-based TTS with Merlin. The performance is shown in Table~\ref{tbl:result_baseline}.
	\begin{table}[h]
		 
		\centering
		\caption{Official ZeroSpeech 2019 baseline and topline result.}
		\label{tbl:result_baseline}
		\begin{tabular}{|c|c|c|}
			\hline
			\multicolumn{1}{|c|}{\textbf{Feature}} & \multicolumn{1}{c|}{\textbf{ABX}} & \multicolumn{1}{c|}{\textbf{Bit rate}} \\ \hline
			Baseline & 35.63 & 71.98 \\ \hline
			Topline & 29.85	& 37.73 \\ \hline
		\end{tabular}
	\end{table}

	\subsection{Preliminary model}
	We started to explore this challenge using a simpler method and gradually increased our model’s complexity.
	\subsubsection{Direct feature representation}
	We directly evaluated the ABX and the bit rate of Mel-spectrogram and MFCC as speech representations. In Table~\ref{tbl:result_direct}, we report each feature extraction method with respect to their ABX and bit rates. In our preliminary experiments, MFCC produced better performances on the ABX metric than the Mel-spectrogram. Therefore, for the rest of our discussion, we only focus on utilizing MFCC features. However, even the MFCC has better ABX score, the bit rate still remains too high.
	\begin{table}[h]
		\centering
		\caption{Direct feature representation (MFCC and Mel-spec) result on ABX with DTW cosine distance and bit rate.}
		 
		\label{tbl:result_direct}
		\begin{tabular}{|c|c|c|}
			\hline
			\multicolumn{1}{|c|}{\textbf{Feature}} & \multicolumn{1}{c|}{\textbf{ABX}} & \multicolumn{1}{c|}{\textbf{Bit rate}} \\ \hline
			Mel-Spec & 30.291 & 1738.38 \\ \hline
			MFCC & 21.114 & 1737.47 \\ \hline
		\end{tabular}
		 
	\end{table}

	\subsubsection{K-Means}
	We trained Minibatch K-Means (with scikit-learn toolkit \cite{scikit-learn}) on the MFCC feature and varied the cluster size: 64, 128, 256. We represent a data point (a speech frame) K-Means by using the closest centroid vector to the data frame and calculate the ABX with the DTW cosine.
	Table~\ref{tbl:result_kmeans} reports all the models and their configurations with respect to their ABX and bit rate.
	% \usepackage{multirow}
	\begin{table*}[h]
		\caption{K-Means continuous representation result on ABX and bit rate. C is codebook size, T is time reduction.}
		\centering
		\label{tbl:result_kmeans}
		\begin{tabular}{|c|c|c|c|c|}
			\hline
			\textbf{Model} & \multicolumn{4}{c|}{\textbf{ABX / Bitrate}} \\ \hline
			\multirow{4}{*}{\begin{tabular}[c]{@{}c@{}}K-Means\\ (cont, \\ DTW cos)\end{tabular}} & \textbf{\#C} & \textbf{1T} & \textbf{2T} & \textbf{4T} \\ \cline{2-5} 
			& \textbf{64} & 23.56 / 553 & 25.97 / 280 & 29.41 / 136 \\ \cline{2-5} 
			& \textbf{128} & 23.16 / 649 & 24.24 / 321 & 28.12 / 161 \\ \cline{2-5} 
			& \textbf{256} & 21.90 / 744 & 23.73 / 369 & 27.17 / 182 \\ \hline
		\end{tabular}
	\end{table*}

	\subsubsection{Gaussian Mixture Model (GMM)}
	We trained GMM with diagonal covariance matrices (with scikit-learn toolkit \cite{scikit-learn}) on the MFCC features. We varied the number of mixtures: 64, 128, and 256. We represent a data point (a speech frame) with the posterior probability from each component with a Bayes rule $p(z|x) \propto p(x|z) p(z)$ and calculate the ABX with DTW KL-divergence.
	In Table~\ref{tbl:result_gmm}, we report all of the models and their configurations with respect to their ABX and bit rate.
	% \usepackage{multirow}
	\begin{table*}[h]
		\caption{GMM posterior representation result on ABX and bit rate. C is codebook size, T is time reduction}
		\centering
		\label{tbl:result_gmm}
		\begin{tabular}{|c|c|c|c|c|}
			\hline
			\textbf{Model} & \multicolumn{4}{c|}{\textbf{ABX / Bit rate}} \\ \hline
			\multirow{4}{*}{\begin{tabular}[c]{@{}c@{}}GMM\\ (post,\\ DTW KL)\end{tabular}} & \textbf{\#C} & \textbf{1T} & \textbf{2T} & \textbf{4T} \\ \cline{2-5} 
			& \textbf{64} & 20.81 / 1647 & 22.67 / 676 & 29.82 / 257 \\ \cline{2-5} 
			& \textbf{128} & 19.61 / 1705 & 23.06 / 704 & 31.19 / 281 \\ \cline{2-5} 
			& \textbf{256} & 18.93 / 1691 & 23.39 / 757 & 32.99 / 306 \\ \hline
		\end{tabular}
	\end{table*}
	
	\subsection{Proposed model}
	
	% \usepackage{multirow}
	\begin{table*}[h]
		\centering
		\caption{VQ-VAE codebook representation result on ABX and bit rate.  C is codebook size, T is time reduction. Blue font denotes our submitted system.}
		 
		\label{tbl:result_vqvae}
		\begin{tabular}{|c|c|c|c|c|c|}
			\hline
			\textbf{Model} & \multicolumn{5}{c|}{\textbf{ABX / Bit rate}} \\ \hline
			& \textbf{\#CL} & \textbf{1T} & \textbf{2T} & \textbf{4T} & \textbf{8T} \\ \cline{2-6} 
			& \textbf{64} & 27.46 / 606 & 25.51 / 302 & 26.15 / 138 & 28.81 / 70 \\ \cline{2-6} 
			& \textbf{128} & 27.65 / 686 & 24.29 / 347 & 25.04 / 165 & 30.87 / 79 \\ \cline{2-6} 
			& \textbf{256} & 27.63 / 787 & {\color[HTML]{3166FF} 24.37 / 349} & {\color[HTML]{3166FF} 24.17 / 184} & 30.51 / 79 \\ \cline{2-6} 
			\multirow{-5}{*}{\begin{tabular}[c]{@{}c@{}}VQ-VAE\\ (cont, \\ DTW cos)\end{tabular}} & \textbf{512} & 27.69 / 871 & 23.59 / 400 & 24.63 / 180 & 32.02 / 74 \\ \hline
		\end{tabular}
		 
	\end{table*}

	\subsubsection{VQ-VAE}
	Next we describe our encoder and decoder architecture in Fig.~\ref{fig:vqvae_4t_model} with four times the sequence length reduction. For the input and output targets, we use the MFCC features and explore different stride sizes to reduce the time length from 1, 2, 4, 8. We use speaker embedding with 32 dimensions and codebook embedding with 64 dimensions. We varied the number of codebooks: 64, 128, 256, 512. Batch normalization \cite{ioffe2015batch} and LeakyReLU \cite{maas2013rectifier} activation were applied to every layer, except the last encoder and decoder layer. The decoder input is a concatenation between codebook and speaker embedding in the channel axis. We set commitment loss coefficient $\gamma=0.25$.

	\subsubsection{Multi-scale Code2Spec inverter}
	 
\begin{figure}[]
		\centering
		\includegraphics[width=0.95\linewidth]{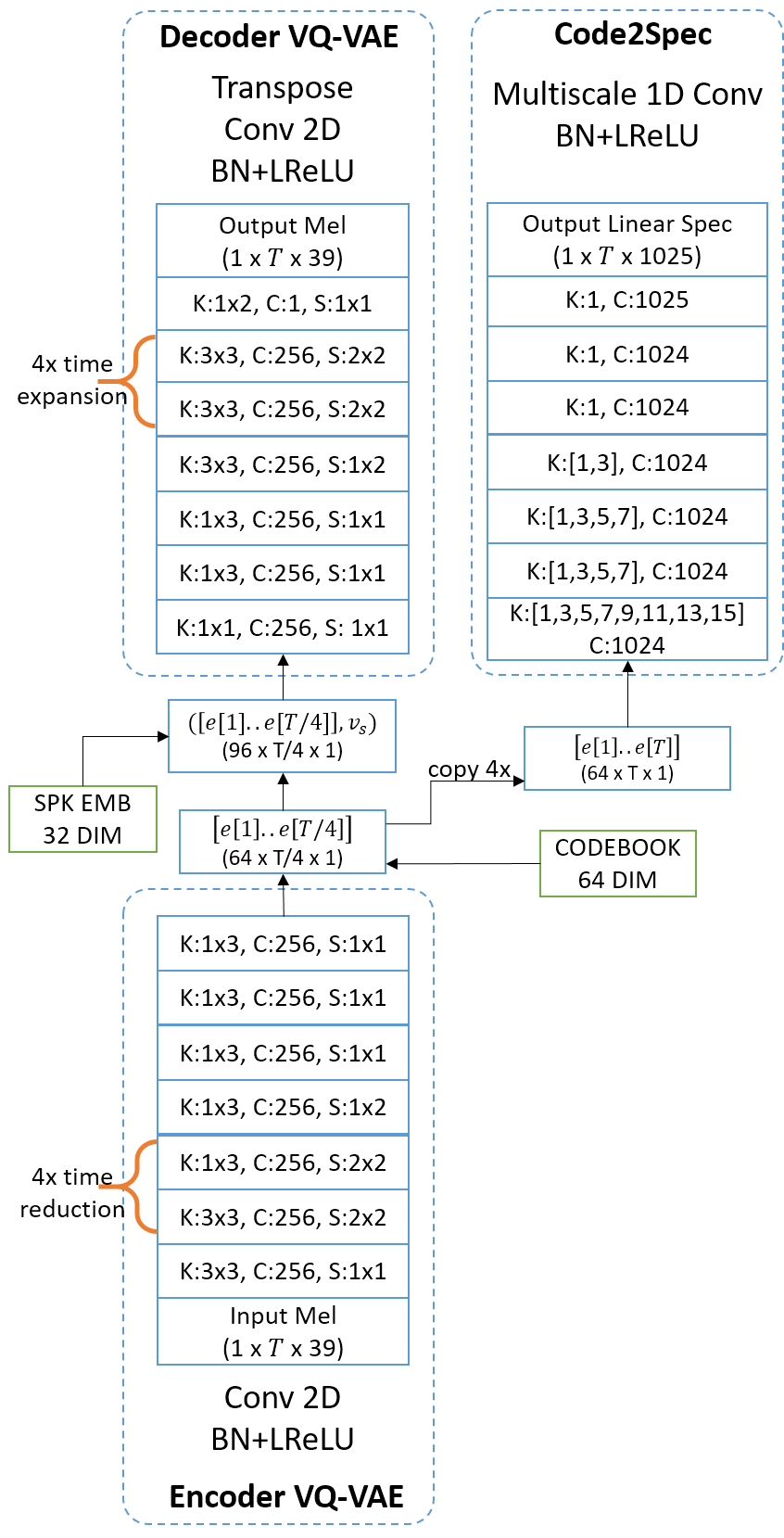}
		 
		\caption{Left: VQ-VAE encoder  and decoder architecture with 4x time reduction (based on stride size in encoder layer). Right: Code2Spec architecture. Definition: K is \textbf{k}ernel size, C is output \textbf{c}hannel, S is \textbf{s}tride size, and T is input frame length. \texttt{K:3x3} denotes 2D convolution with 3x3 kernel size across time and frequency axis, \texttt{K:[1,3,5,7]} denotes 1D convolution with 4 different kernel size (1, 3, 5, 7) across time-axis.}
		\label{fig:vqvae_4t_model}
	\end{figure}

	In Fig.~\ref{fig:vqvae_4t_model}, we describe our inverter architecture. Our input is a codebook sequence with 64 dimensions and our target output is a sequence of linear magnitude spectrogram with 1025 dimensions. The first four layers have multiple kernels with different sizes across the time-axis. All convolution layers have stride = 1 and the ``same'' padding. Batch normalization and LeakyReLU activation are applied to every layer, except the last one before the output prediction. For the adversarial loss, we found LSGAN is stabler, thus LSGAN with $\beta=1$ is used in every model. We independently trained the inverter to generate a voice target speaker with a \texttt{train/voice} set. We have two inverters for the English set and one for the surprise set.

	\subsubsection{Model training}	 
	
	We used Adam \cite{kingma2014adam} as our first-order optimizer for both VQ-VAE and the Code2Spec inverter. All of our models are implemented with PyTorch \cite{paszke2017automatic} framework.
	
	\subsubsection{Results and Discussion}

	%Table~\ref{tbl:result_vqvae} reports all models and their configurations with respect to their ABX and bit rate. We also attempted further enhancements of the synthesized output using several techniques, such as Wavenet \cite{oord2016wavenet} and GAN voice conversion \cite{sisman2018adaptive}. Unfortunately, we failed to further improve the performance. After considering the balance between the discrimination score ABX and the bit-rate compression rate, we submitted two proposed systems: (1) 256 codebooks and 4 stride size to reduce the time length and (2) 256 codebooks and 2 stride size  to reduce the time length.
	
	Table~\ref{tbl:result_vqvae} reports all models and their configurations with respect to their ABX and bit rate. Considering the balance between the discrimination score ABX and the bit-rate compression rate, we submitted two proposed systems: (1) 256 codebooks and 4 stride size to reduce the time length and (2) 256 codebooks and 2 stride size  to reduce the time length. 
	
	We also attempted further enhancement of the synthesized voice using several techniques, such as WaveNet \cite{oord2016wavenet, chorowski2019unsupervised} and GAN-based voice conversion \cite{sisman2018adaptive}. WaveNet decoder is conditioned by frame-wise linguistic features or acoustic features with a 5ms timeshift (80 times smaller than the speech samples). As the sample rate of the codebook embeddings of our system was 320 times smaller than the speech samples, the Wavenet couldn't produced satisfying result. GANs are known to be effective for achieving high-quality voice conversion with clean input data \cite{fang2018high, hsu2017voice}. However, our task is more challenging due to the fact that our generated voice will always have some distortion. Therefore, GAN-based voice conversion approach failed to improve our performance. 
	As a future work, we will investigate the use of GAN-based speech enhancement \cite{meng2018cycle} approaches to further improve our results.
	 
	\section{Conclusions}
	 
	We described our approach for the ZeroSpeech Challenge 2019 for unsupervised unit discovery. We explored many different possibilities: feature extraction, clustering algorithm, and embedding representation. For our final submission, we utilized VQ-VAE to extract a sequence of codebook vectors. The codebook generated by VQ-VAE has a better trade-off between ABX and the bit rate compared to the other models such as K-Means, GMM, or direct feature representation. To reconstruct speech from the codebook, we trained a Code2Spec inverter to generate a corresponding linear magnitude spectrogram. The combination between VQ-VAE and Code2Spec significantly improved the intelligibility (in CER), the MOS, and the discrimination ABX scores compared to the official ZeroSpeech 2019 baseline or even the topline.

	\section{Acknowledgements}
	 
	Part of this work was supported by JSPS KAKENHI Grant Numbers JP17H06101 and JP17K00237.
	\bibliographystyle{ieeetr}
	\bibliography{refs}
\end{document}